\documentclass{article}
\usepackage{ijcai23}

\usepackage{times}
\usepackage{soul}
\usepackage{url}
\usepackage[hidelinks]{hyperref}
\usepackage[utf8]{inputenc}
\usepackage[small]{caption}
\usepackage{graphicx}
\usepackage{amsmath}
\usepackage{amsthm}
\usepackage{booktabs}
\usepackage{algorithm}
\usepackage{algorithmic}
\usepackage[switch]{lineno}
\usepackage[square,sort&compress,numbers]{natbib}
\usepackage{color}
\usepackage{xcolor}
\usepackage{colortbl}
\definecolor{mygray}{gray}{.9}
\definecolor{mypink}{rgb}{.99,.91,.95}
\definecolor{mycyan}{cmyk}{.3,0,0,0}
\usepackage{stfloats}
\usepackage{multirow}
\title{FinVis-GPT: A Multimodal Large Language Model for Financial Chart Analysis}

\author{Anonymous submission}
\author{
Ziao Wang
\and
Yuhang Li\and
Junda Wu\and
Jaehyeon Soon\and
Xiaofeng Zhang\footnote{Corresponding author. Email: zhangxiaofeng@hit.edu.cn}\and
\affiliations
School of Computer Science and Technology, Harbin Institute of Technology, Shenzhen, China\\
\emails
wangziao1993@hotmail.com, 331137797@qq.com, wujunda@stu.hit.edu.cn, jaehyeon\_soon@stu.hit.edu.cn, zhangxiaofeng@hit.edu.cn
}

\begin{document}

\maketitle

\begin{abstract}
In this paper, we propose FinVis-GPT, a novel multimodal large language model (LLM) specifically designed for financial chart analysis. By leveraging the power of LLMs and incorporating instruction tuning and multimodal capabilities, FinVis-GPT is capable of interpreting financial charts and providing valuable analysis. To train FinVis-GPT, a financial task oriented dataset was generated for pre-training alignment and instruction tuning, comprising various types of financial charts and their corresponding descriptions. We evaluate the model performance via several case studies due to the time limit, and the promising results demonstrated that FinVis-GPT is superior in various financial chart related tasks, including generating descriptions, answering questions and predicting future market trends, surpassing existing state-of-the-art multimodal LLMs. The proposed FinVis-GPT serves as a pioneering effort in utilizing multimodal LLMs in the finance domain and our generated dataset will be release for public use in the near future to speedup related research. 
\end{abstract}

\section{Introduction}
In the era of large language model (LLM) \cite{wei:llama:2023,selfinstruct:wang:2022,chatgpt,openai2023gpt4}, various real-world applications will be deeply and permanently changed by the LLMs as well as other large models (LMs). %The rapid advancement of artificial intelligence and machine learning technologies has revolutionized numerous fields, including finance. 
For instance, the LLMs already demonstrated a superior performance in various NLP tasks such as understanding and generating human-like text. Similarly, the large multimodal models (LMMs) %taking  take multimodal inputs, such as text and images, 
has opened up new possibilities for more complex applications such as embodied robot. Thus, a good number of research efforts as well as industrial attentions have been attracted to explore the possibility whether such LMs could be utilized for financial related tasks. %the possibilities that these LMs However, the application of these multimodal LLMs in the financial domain remains largely unexplored.

Therefore, we are motivated to propose this novel multimodal large language model (FinVis-GPT) specifically designed for understanding financial chart. The proposed approach are two-stage ones. At the first stage, we must carefully prepare a dataset for this task which will be released for public use in the near future. At the second stage, we train a large multimodal model using this dataset. Note that it is very demanding to tune a large multimodal model from the begining. Thus, we only fine-tune an existing model using this generated dataset. We expect that, by leveraging the power of LLMs, the proposed FinVis-GPT should be capable of interpreting financial charts and providing more accurate analysis in a human-like manner. This capability allows FinVis-GPT to answer a wide range of questions, such as predicting future trends based on historical data, identifying key patterns, and providing explanations for observed market phenomena.
%---------------------------------
\begin{figure*}[t]
  \centering
  \includegraphics[width=0.99\linewidth]{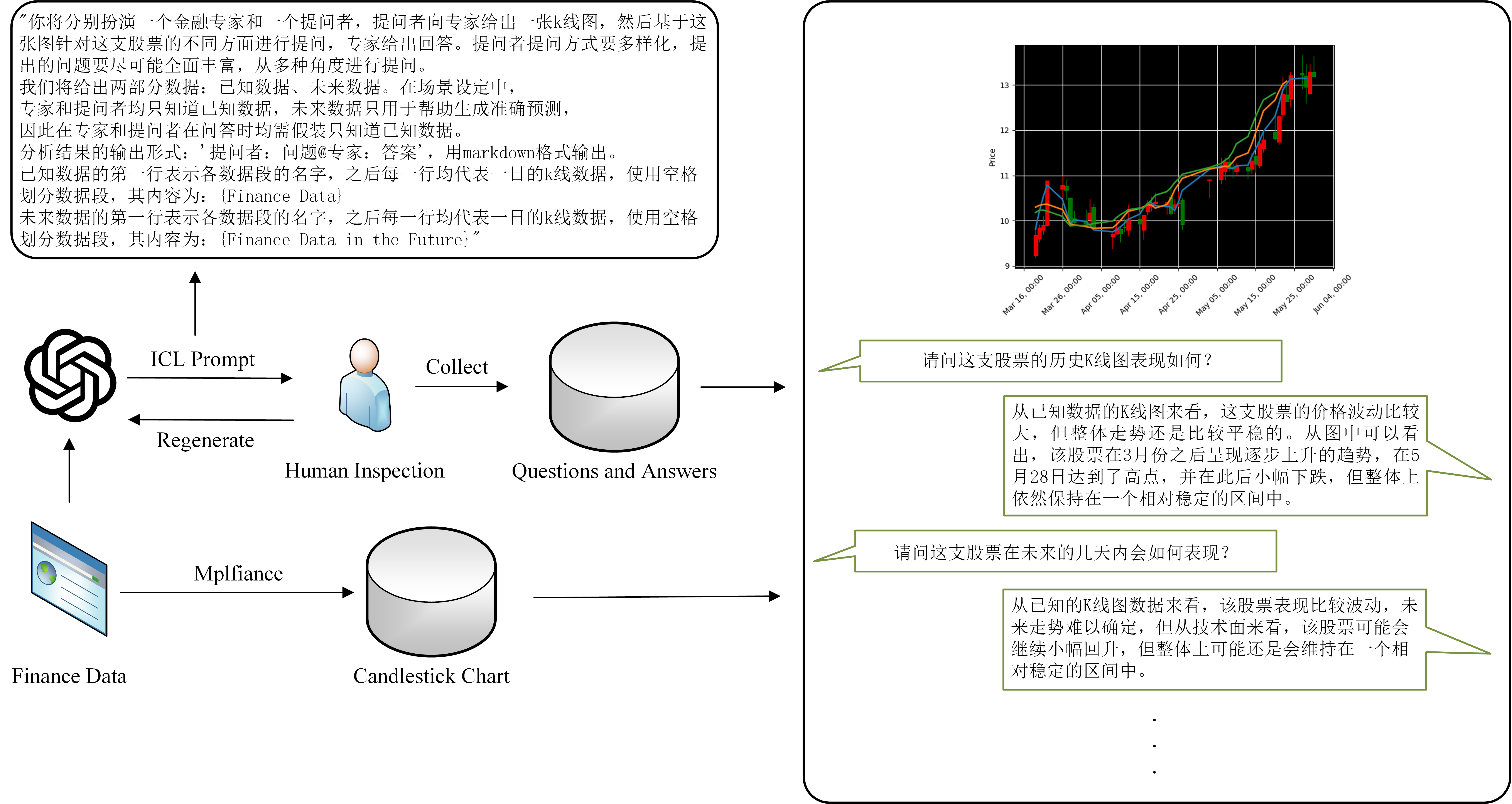}
  \caption{The designed process to generate multimodal dataset.}\label{fig:example}
\end{figure*}
%---------------------------------

As aforementioned, the key contribution of our work is the creation of a financial task oriented dataset for pre-training and instruction-tuning the large models. For the pre-training phase, we have curated a dataset comprising various types of financial charts along with their corresponding descriptions. This dataset enables FinVis-GPT to learn the intricate relationships between visual patterns in financial charts and their textual interpretations. For the instruction tuning phase, we have prepared a dataset that pairs images of financial charts with a set of instructions or questions. This dataset allows FinVis-GPT to learn how to respond to specific queries related to financial chart analysis, thereby enhancing its ability to generate relevant and accurate responses. After training the 
FinVis-GPT on this dataset, we investigate the model performance via various case studies due to the time limit. The results demonstrated that FinVis-GPT can effectively analyze financial charts and generate reliable and accurate interpretations. We believe that our work paves the way for more sophisticated applications of multimodal LLMs in the financial domain, potentially transforming how financial analysis is conducted. 

\section{Related Work}
The evolution of LLMs and LMMs have already become the major research subjects recently. In this section, we briefly review several most pertinent works in these areas and discuss their relationship to our proposed model, FinVis-GPT.

\paragraph{Large Language Models and Instruction Tuning} The transformation of LLMs into instruction followers has been a prominent research direction. For instance, InstructGPT \cite{ouyang:instructgpt:2022} was introduced as a model designed to follow instructions given in natural language and generate useful responses. This model demonstrated that instruction tuning could significantly enhance the performance of LLMs, surpassing even the capabilities of GPT-3. Building on this concept, \citet{vicuna2023} fine-tuned the LLaMA model \cite{wei:llama:2023} on user-shared dialogues collected from ShareGPT, resulting in an open-source chatbot with impressive performance.

\paragraph{Large Multimodal Models} The extension of LLMs to handle multimodal inputs has been a significant advancement in recent research. The KOSMOS-1 model \cite{huang:kosmos:2023}, trained from scratch on web-scale multimodal corpora, showcased impressive performance across language understanding, generation, and perception-language tasks. Similarly, MiniGPT-4 \cite{zhu:MiniGPT4EnhancingVisionLanguage:2023} demonstrated the potential of aligning a frozen visual encoder with a frozen LLM, Vicuna, using a single projection layer. Further extending the multimodal capabilities, mPLUG-Owl \cite{ye:MPLUGOwlModularizationEmpowers:2023} was proposed to concurrently support multiple modalities and facilitate diverse unimodal and multimodal abilities through modality collaboration. In a similar vein, LLaMA-Adapter V2 \cite{gao:LLaMAAdapterV2ParameterEfficient:2023} was proposed as a parameter-efficient model capable of handling visual instructions. Lastly, InstructBLIP \cite{dai:InstructBLIPGeneralpurposeVisionLanguage:2023} was designed to handle a variety of instructions, showcasing its ability to generate detailed captions, count specific objects, and address general inquiries posed by users.

Building upon these advancements, our proposed model, FinVis-GPT, incorporates financial charts as part of the multimodal input. This integration enables a more nuanced understanding of financial data, marking a significant step towards the application of multimodal LLMs in the financial domain. By leveraging the strengths of both instruction tuning and multimodal capabilities, FinVis-GPT aims to provide insightful analysis of financial charts, demonstrating the potential of multimodal LLMs in domain-specific applications.

\section{Generating Multimodal Financial Dataset}

The data collection for FinVis-GPT involved creating datasets for two phases: pre-training alignment and instruction tuning. The goal of these datasets was to equip the model with the ability to understand and interpret multimodal data, particularly financial charts, and generate valuable responses based on given instructions. An illustrative example of our whole collection pipeline and the collected data is shown in Figure \ref{fig:example}.

\subsection{Pre-training Alignment Dataset}

Pre-training alignment is a crucial step in training multimodal models, as it allows the model to align various types of data into a common embedding space. For the purpose of this step, we used historical daily stock price data of Chinese A-share from 2006 to 2023. This data was segmented into smaller sets containing 60-80 trading days, and each set was further divided into prompt data (data given to the model for prediction) and predict data (data to be predicted), with the former comprising 60-80\% of each set.

Images were generated from this prompt data using the mplfinance\footnote{https://github.com/matplotlib/mplfinance} library, with a split of 80\% for candlestick charts and 20\% for line charts. To simulate real world scenarios, the generated charts were enhanced with moving averages of 3, 6, and 9 days, volume bars, and various chart styles, all added randomly.

The data structure for each entry in this dataset consisted of an image, an instruction, and an answer. The instructions, designed to request an interpretation of the charts, were manually crafted. The answer for each instruction was generated by using chatGPT to interpret the prompt data. The prompt given to chatGPT are shown in Table \ref{tab:pre-train-prompt}.

% \usepackage{booktabs}
\begin{table}[]
\centering
\small
\begin{tabular}{p{8cm}}
\toprule[2pt]
\\
You will play the role of a financial expert. Upon receiving a k-line chart, you should first identify what type of chart it is and then describe the different stages of stock trends. You are required to conduct professional financial analysis on the input data while ensuring that your analysis is comprehensive and professional from different perspectives. Finally, you need to summarize your findings. To facilitate generating answers, you will not receive an image but rather data related to the k-line chart. In this scenario, since it is assumed that you are analyzing an image as an expert, your answer should pretend that you are analyzing an image and only mention content commonly found in k-line charts.

In your answer:
\begin{itemize}
    \item Do not evaluate what you are doing; simply provide answers.
    \item Use "this stock" instead of direct stock codes.
    \item Do not explain the meaning of different data segments or their names.
    \item Do not draw charts; use text descriptions based on data only.
    \item Avoid saying more data is needed or suggesting other factors be considered; provide clear analytical conclusions instead.
\end{itemize}
The output format for analysis results: `Answer', using markdown format. The first line of received content represents the name of each data segment, with each subsequent line representing one day's k-line data separated by spaces.
 
\\
\bottomrule[2pt]
\end{tabular}
\caption{Prompt designed for pre-training stage in data collection.}
\label{tab:pre-train-prompt}
\end{table}
%------------------------------------------------------
\subsection{Instruction Tuning Dataset}

For instruction tuning, a separate dataset was created, comprising 200K sets, each with about five questions. The primary purpose of this dataset was to fine-tune FinVis-GPT's ability to generate relevant and accurate responses to specific queries related to financial chart analysis.

Like the pre-training alignment dataset, the data structure for this dataset also consists of an image, an instruction, and an answer. However, the key difference lies in the generation manner of instructions and answers: they were crafted using both prompt data and predict data, which allowed the model to generate future trend predictions, the prompt designed to give chatGPT is reported in Table \ref{tab:instruc-prompt}. 

% \usepackage{booktabs}
\begin{table}[]
\centering
\small
\begin{tabular}{p{8cm}}
\toprule[2pt]
\\

You will play the roles of a financial expert and a questioner respectively. The questioner presents a candlestick chart to the expert, and then asks questions based on different aspects of this stock. A total of five `Q\&A' sessions are required. The questioning should be diverse, with questions as comprehensive and rich as possible, from multiple perspectives. For example, asking the expert for professional financial analysis of the stock, predicting its future trend or explaining data. To facilitate prediction, we will provide two sets of data: known data and future data. In this scenario, it is assumed that both the expert and questioner only know about the known data; while future data is only used to help generate accurate predictions. Therefore, when answering questions related to predictions, experts must pretend that they only know about known data without mentioning future data explicitly but must combine them with generated predictions using future data to answer such questions clearly.

In addition, since you are playing an expert who receives images in this scenario; your answers need to use language suitable for analyzing an image-oriented perspective while focusing on content commonly found in candlestick charts. During questioning by the questioner or answering by the expert:
\begin{itemize}
    \item Do not directly use stock codes; instead use `this stock' as a substitute.
    \item Do not explain what each segment represents.
    \item Do not mention any segment names.
    \item Experts should refrain from evaluating their own tasks during responses.
    \item Avoid saying more information is needed or suggesting other factors need consideration.
    \item Do not allow questioner to view additional information themselves.
\end{itemize}

When asked for predictions by questioner during questioning sessions - experts must give out forecasts based on generated results using future-data. The output format for analytical results: "Question@Answer@" where both "@" symbols serve as separators between Question and Answer pairs except at any other time within either part (use markdown formatting). 
The first line of the known data indicates each segment's name followed by k-line trading records for each day, with data segments separated by spaces:

$\{prompt\_data\}$

The first line of the future data indicates each segment's name followed by k-line trading records for each day, with data segments separated by spaces:

$\{predict\_data\}$

\\
\bottomrule[2pt]
\end{tabular}
\caption{Prompt designed for instruction-tuning stage in data collection.}
\label{tab:instruc-prompt}
\end{table}
%------------------------------------------------------
\subsection{Dataset Statistics}
Table \ref{tab:statisc} provides a detailed breakdown of the key statistics associated with the collected datasets used in the pre-training and instruction tuning phases of the FinVis-GPT model. The count of words in questions, answers, and total dialog exchanges (denoted as `\#') are examined under various statistical metrics such as mean, the 5-th percentile (q-5\%), and the 95th percentile (q-95\%).

During pre-training, we observe that on average, questions have around 28.68 words, while answers contain approximately 401.15 words. This indicates that responses tend to be much more detailed and comprehensive. The entire dialog, including both questions and answers, contains about 429.83 words on average. The data distributions for the number of words in the questions, answers, and the entire dialog show a wide spread, as evidenced by the 5th and 95th percentile values.

In the instruction tuning phase, the number of turns taken averages at 4.79, hinting at the complexity and depth of the conversations in the dataset. Questions contain fewer words compared to the pre-training dataset, with an average of 19.96 words. The answers in this phase are significantly shorter, with approximately 63.03 words on average. This suggests a shift towards more focused and concise communication. The entire dialog contains about 397.36 words on average, with a less pronounced spread than observed in the pre-training dataset.
% \usepackage{booktabs}
% \usepackage{multirow}
\begin{table}[]
\centering
\begin{tabular}{@{}clccc@{}}
\toprule[1.5pt]
\multicolumn{1}{l}{}         &                    & mean   & q-5\% & q-95\% \\ \midrule
\multirow{3}{*}{pre-train}   & \# Question & 28.68  & 21          & 36           \\
                             & \# Answer   & 401.15 & 179         & 882          \\
                             & \# Dialog & 429.83 & 207         & 910          \\ \midrule
\multirow{4}{*}{instruction} & \# Turns     & 4.79   & 3.00        & 5.00         \\
                             & \# Question & 19.96  & 11.00       & 14.00        \\
                             & \# Answer   & 63.03  & 23          & 41           \\
                             & \# Dialog & 397.36 & 238         & 748          \\ \bottomrule[1.5pt]
\end{tabular}
\caption{Summary statistics of the collected dataset. Here, `\#' represents word count. The dataset includes two main categories: pre-training and instruction. The statistics cover the mean, 5th percentile (q-5\%) and 95th percentile (q-95\%) of word count for questions, answers, and dialogues in each category.}

\label{tab:statisc}
\end{table}
%------------------------------------------
%---------------------------------
\begin{figure}[]
  \centering
  \includegraphics[width=0.99\linewidth]{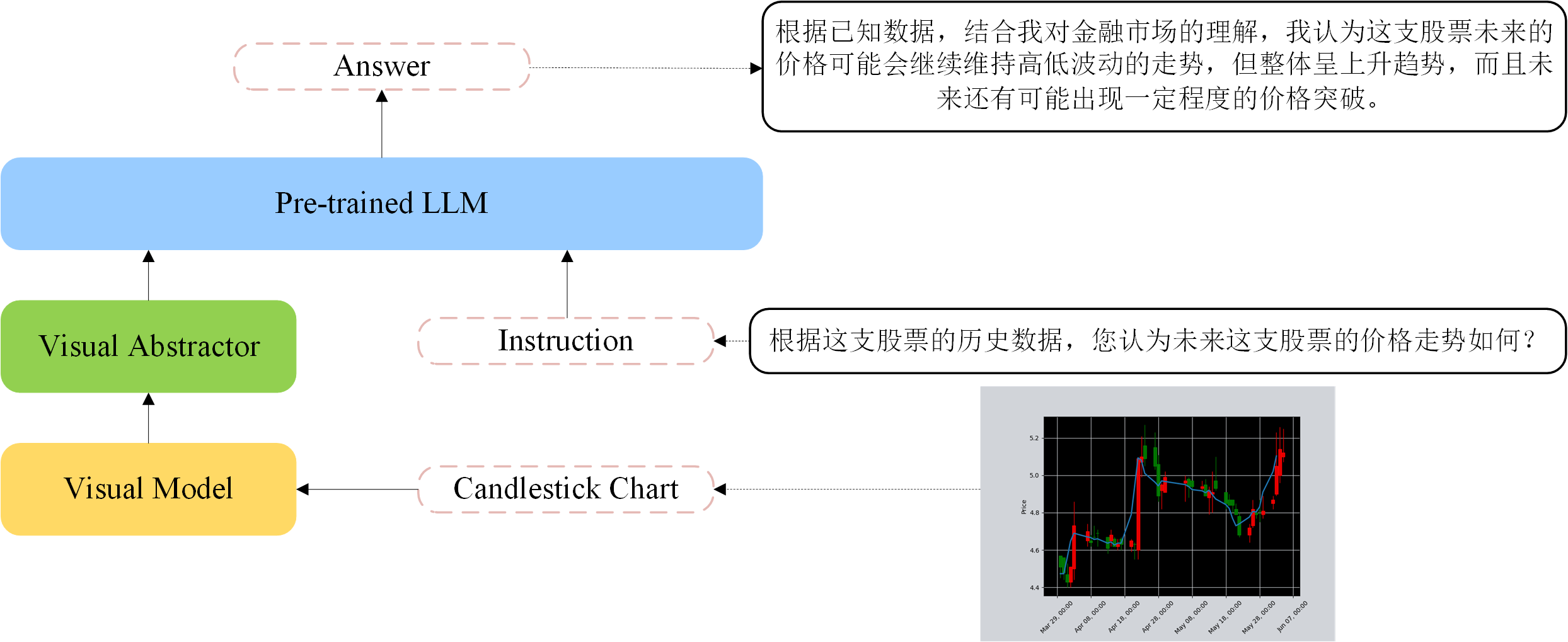}
  \caption{The model architecture.
  }\label{fig:model}
\end{figure}
%---------------------------------
\section{Model Training}
The FinVis-GPT model was built on top of the pre-existing LLaVA \cite{liu:llava:2023} model, incorporating the advanced language capabilities of the latter while extending them for the specific financial context, the model architecture is plotted in Figure \ref{fig:model}. The training process consists of two major steps: pre-training alignment and instruction tuning.
%---------------------------------
\begin{figure*}[]
  \centering
  \includegraphics[width=0.99\linewidth]{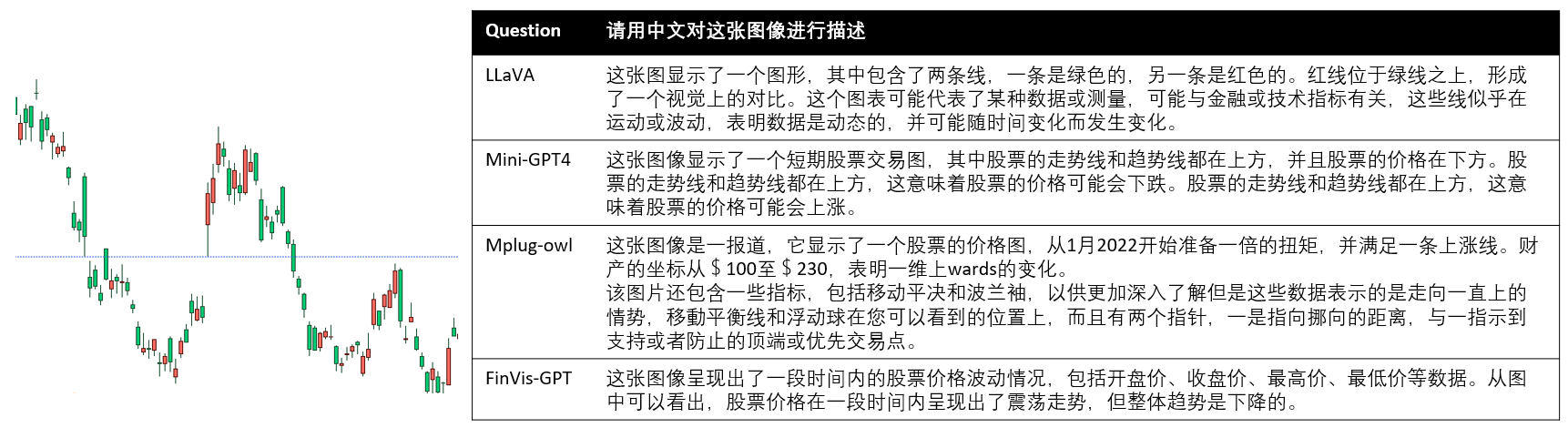}
  \caption{Experiment results on description generation.
  }\label{fig:descrition}
\end{figure*}
%---------------------------------
%---------------------------------
\begin{figure*}[]
  \centering
  \includegraphics[width=0.99\linewidth]{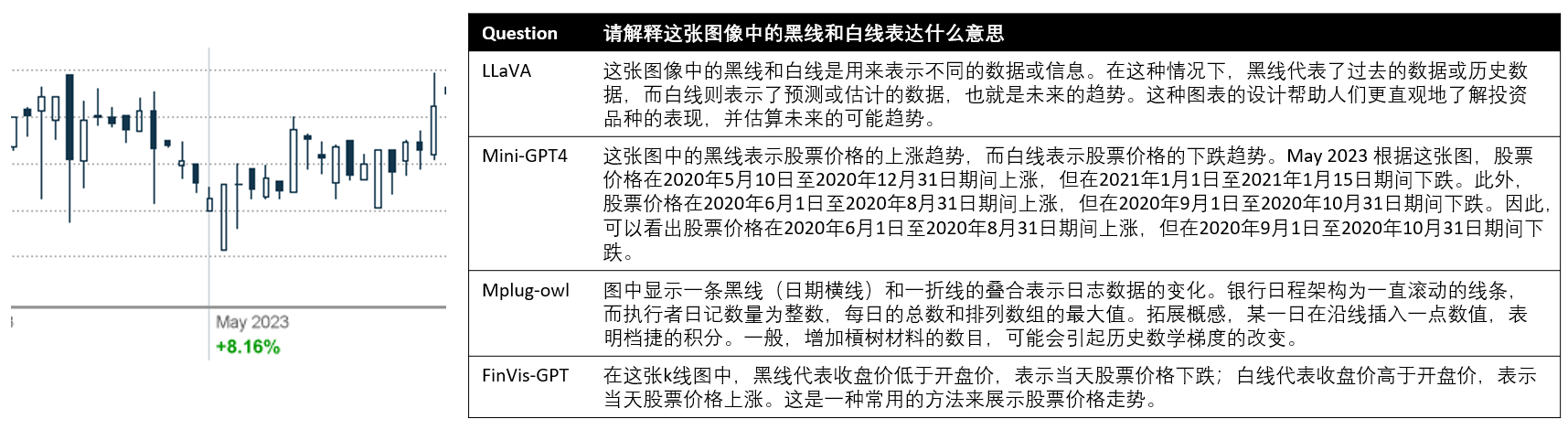}
  \caption{Experiment results on question answering.
  }\label{fig:qa}
\end{figure*}
%---------------------------------
%---------------------------------
\begin{figure*}[]
  \centering
  \includegraphics[width=0.99\linewidth]{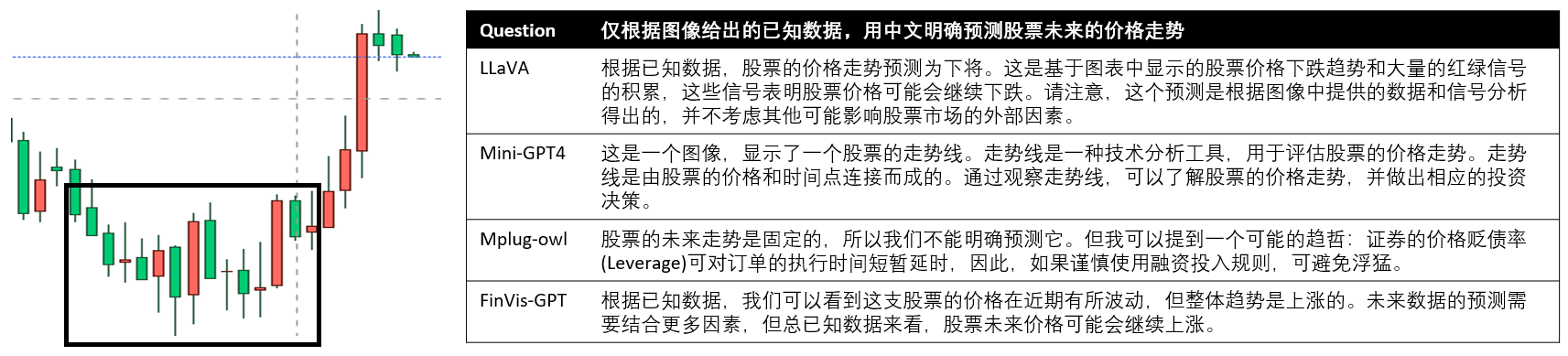}
  \caption{Experiment results on trend prediction.
  }\label{fig:predict}
\end{figure*}
%---------------------------------

\subsection{Pre-training Alignment}

Pre-training alignment aimed at teaching the model to understand the relationship between visual patterns in financial charts and their corresponding textual interpretations. The pre-training alignment dataset, consisting of various financial charts and corresponding descriptions, was used for this purpose.

For the pre-training, we adopted the same training approach as LLaVA but used our specifically curated dataset of financial charts and descriptions. The model was trained using a batch size of 128 and a learning rate of 2e-3. The pre-training was carried out on 8 NVIDIA Tesla A100 GPUs for 1 epochs.

The effectiveness of pre-training alignment was evaluated by feeding the model with new, unseen financial charts and checking its ability to generate accurate and relevant descriptions. The generated descriptions were evaluated by a panel of financial experts for their accuracy and relevance. %We show an example of descriptions in Figure \ref{fig:des}.
% %---------------------------------
% \begin{figure}[!t]
%   \centering
%   \includegraphics[width=0.99\linewidth]{decription.png}
%   \caption{An example of chart descriptions.
%   }\label{fig:des}
% \end{figure}
% %---------------------------------
\subsection{Instruction Tuning}

Instruction tuning is a technique that allows the model to learn how to generate appropriate responses to specific instructions or queries. For this, we used the instruction tuning dataset, which was specifically created for the purpose of fine-tuning FinVis-GPT.

The tuning phase involved adjusting the model's parameters so that it could accurately respond to instructions about financial charts. This phase was also executed using a batch size of 128 and a learning rate of 1e-5 for 3 epochs.

\subsection{Regularization and Model Validation}

To prevent overfitting during the training process, we incorporated dropout and weight decay regularization techniques. We also used early stopping based on the validation set performance to determine the optimal number of training epochs.

Model validation was performed intermittently throughout the training process. We maintained a holdout validation set that was not used during the training process. At the end of each epoch, the model was tested on this validation set to gauge its performance and to ensure it was learning the intended tasks effectively.

In sum, the training process of FinVis-GPT was a meticulous process aimed at harnessing the language prowess of LLaVA and tailoring it to the complex task of financial chart interpretation and analysis.

\section{Experiments}
\subsection{Experimental Setup}

We compared FinVis-GPT against several baseline models including LLaVA \cite{liu:llava:2023}, MPLUG-Owl \cite{ye:MPLUGOwlModularizationEmpowers:2023}, and MiniGPT-4 \cite{zhu:MiniGPT4EnhancingVisionLanguage:2023}. Each of these models represents the latest advancements in multimodal learning with unique advantages. The metrics used for comparison included quality of financial chart descriptions, understanding of financial context, and prediction accuracy of financial trends. We employed the following three tasks to evaluate each model:

\begin{itemize}
\item \textbf{Description Generation:} For this task, the models were given an image of a financial chart and were required to generate a description, capturing the key trends, patterns, and anomalies. 
\item \textbf{Question Answering:} This task involved a comprehension test where models were given an image of a financial chart along with a set of questions. The questions were designed to assess the model's understanding of the financial context of the chart. 
\item \textbf{Trend Prediction:} For this task, models were provided an image of a financial chart along with historical financial data and were asked to predict future trends. The predictions were compared with actual future data to evaluate the model's predictive performance.
\end{itemize}

\subsection{Results and Discussion}
\paragraph{Description Generation.} The task of description generation is exemplified in Figure \ref{fig:descrition}, where a randomly selected outcome is presented. Based on these results, it is obvious that LLaVA fails to accurately identify the image as a representation of stock trends. In contrast, Minit-GPT4 demonstrated a superior understanding by correctly recognizing the image as a stock trading chart, though it inaccurately identified the blue line as a stock trend line. Moreover, mplug-owl managed to acknowledge the image as a stock price chart but it introduced several unrelated elements, causing its description to veer off the accurate interpretation. Among all models assessed, FinVis-GPT emerged as the most proficient, correctly recognizing the image and providing a concise and accurate description. This underscores its capacity for generating superior descriptions when compared to the other models in this specific context.

\paragraph{Question Answering.} The question answering task is plotted in Figure \ref{fig:qa}. The results reveal that LLaVA substantially misconstrued the stock trend, erroneously identifying the black candle line as the past trend and the white as the future trend. Meanwhile, Mini-GPT4 muddled the representation of black and white lines, further compounding its output with a significant amount of irrelevant content. The mplug-owl model exhibited a complete lack of recognition for the image, fabricating an entirely unrelated narrative. In contrast, the response provided by FinVis-GPT was both concise and accurate, earning it the top spot amongst the compared models for this task. Its output underscores the superior efficacy of FinVis-GPT in understanding and accurately answering questions based on the given visual representation.

\paragraph{Trend Prediction.} 
An example of trend prediction is depicted in Figure \ref{fig:predict}. The left image represents a market trend over a certain period, with the trend within the black box provided as input to the models. The accurate prediction for this trend should indicate an upward trajectory. However, LLaVA's prediction was contrary to this, presenting a downward trend instead. Mini-GPT4 failed to answer the prediction question accurately, and instead produced unrelated information, a phenomenon often referred to as `hallucination'. Similarly, mplug-owl's output was also characterized by this `hallucinating' issue. In contrast, FinVis-GPT's prediction was not only accurate but also incorporated a proper description of the trend. This showcases FinVis-GPT's superiority in trend prediction tasks, with an ability to provide both accurate and informative responses. 

% \paragraph{Classification result}

%----------------------------------------
% % Please add the following required packages to your document preamble:
% % \usepackage{booktabs}
% \begin{table}[]
% \centering
% \begin{tabular}{@{}lcccc@{}}
% \toprule
% model      & Accuracy    & Precision    & Recall    & F1    \\ \midrule
% LLaVA      & 0.4    & 0.0    & N/A    & N/A    \\
% mplug-OWL  & 0.1429 & 0.1667 & 0.1667 & 0.1667 \\
% mini-GPT4  & 0.5556 & 1.0    & 0.1667 & 0.2857 \\
% FinVis-GPT & 0.8    & 0.6667 & 1.0    & 0.8    \\ \bottomrule
% \end{tabular}
% \caption{}
% \label{tab:my-table}
% \end{table}
%--------------------------------------------

\section{Conclusion}

In this work, we presented FinVis-GPT, a novel large multimodal model tailored to the financial domain, with a focus on financial chart analysis. Our approach integrated the benefits of pre-trained LLMs with a curated dataset sourced directly from the financial sector. The FinVis-GPT model showed significant improvement over existing models in terms of generating accurate, relevant, and financially styled responses. Through the creation of a robust instruction tuning dataset and case studies, we have demonstrated the potential of multimodal LLMs in the financial sector. This work lays the foundation for more sophisticated applications of AI in finance, potentially transforming the landscape of financial analysis. Future work will focus on further expanding the applicability of FinVis-GPT in more diverse financial scenarios and real-time financial decision-making.

\appendix

\bibliographystyle{plainnat}
\bibliography{ijcai23}

\end{document}